\documentclass[journal,transmag]{IEEEtran}
\usepackage{hyperref,color,breqn,multirow}
\usepackage[top=0.7in, left=0.65in]{geometry}
\usepackage{graphicx}
\usepackage{amsmath,amsfonts,amssymb,amscd,bm}
\usepackage{booktabs}
\usepackage{tikz}
\usetikzlibrary{shapes,shapes.geometric,arrows,positioning,patterns,shapes.arrows,fit}
\usepackage{upgreek}
\usepackage{tuda-pgfplots}
\usepackage{pgfplots}
\usepgfplotslibrary{groupplots,dateplot}
\usepgfplotslibrary{groupplots}
\pgfplotsset{compat=newest}
\DeclareUnicodeCharacter{2212}{−}
\makeatletter
\newcommand*{\rom}[1]{\expandafter\@slowromancap\romannumeral #1@}
\makeatother
\newcommand{\removelatexerror}{\let\@latex@error\@gobble}
\makeatother

\newcommand{\kpi}{y}
\allowdisplaybreaks
\makeatletter
\newcommand\notsotiny{\@setfontsize\notsotiny{6}{7}}
\makeatother

\begin{document}
\title{Variational Autoencoder based Metamodeling for Multi-Objective Topology Optimization of Electrical Machines}
\author{\IEEEauthorblockN{Vivek Parekh\IEEEauthorrefmark{1,2},
Dominik Flore\IEEEauthorrefmark{2}, and
Sebastian Schöps\IEEEauthorrefmark{1},}
\IEEEauthorblockA{\IEEEauthorrefmark{1}Computational Electromagnetics Group, Technical University Darmstadt, 64289 Darmstadt, Germany}
\IEEEauthorblockA{\IEEEauthorrefmark{2}Robert Bosch GmbH, Engineering, Acquisition, Building Set (PS-EM/EAB), 70442 Stuttgart, Germany}
\thanks{Manuscript received xxx y, 20zz; revised xxx yy, 20zz and xxx 1, 20zz; accepted xxx 1, 20zz.Dates will be inserted by IEEE; Corresponding author: V. Parekh (e-mail: Parekhvivek49@gmail.com).}
}
\IEEEtitleabstractindextext{%
\begin{abstract}
Conventional magneto-static finite element analysis of electrical machine design is time-consuming and computationally expensive. Since each machine topology has a distinct set of parameters, design optimization is commonly performed independently. This paper presents a novel method for predicting Key Performance Indicators (KPIs) of differently parameterized electrical machine topologies at the same time by mapping a high dimensional integrated design parameters in a lower dimensional latent space using a variational autoencoder. After training, via a latent space, the decoder and multi-layer neural network will function as meta-models for sampling new designs and predicting associated KPIs, respectively. This enables parameter-based concurrent multi-topology optimization. 
\end{abstract}

}
\maketitle
\thispagestyle{empty}
\pagestyle{empty}
\section{Introduction}\label{sec:INT}
\IEEEPARstart{T}{he} PERFORMANCE of an electrical machine is measured by its Key Performance Indicators (KPIs), e.g, torque, power, cost, etc., which are determined in early design stages by using computationally intensive magneto-static finite element (FE) simulations. In \cite{9333549}, it is demonstrated how efficiently cross-domain KPIs are learned and predicted for various machine design representations using image and parameter-based deep learning (DL). It was observed that the method based on parameters results in higher prediction accuracy than the image-based representation. Since the parameter-based approach cannot deal with multiple topologies simultaneously, optimization for various topologies at the same time can not be based on such parameter-based meta models. 
On the other hand, one could use an image-based approach but at the cost of lower prediction accuracy and increased effort due to the generation and processing of the (high-resolution) images.

In this contribution, we optimize differently parameterized topologies of permanent magnet synchronous machine (PMSM) by mapping the high-dimensional combined input design space into a lower-dimensional unified latent space using a Variational Autoencoder (VAE). The VAE is a probabilistic method for transforming high-dimensional input data into a latent space (encoding). It reconstructs input data and allows the generation of new samples in the original design space (decoding) \cite{MAL-056}. While it is typically used for image-based generative modeling, e.g., \cite{NIPS2016_eb86d510} and 
recently in the context of a design problem for an electromagnetic die press \cite{electronics10182185}, we use it for a parameter-based case due to its higher prediction accuracy for the electrical machine KPIs~\cite{9333549}.

The VAE is trained to find a latent space while training a multi-layer perceptron (MLP) for KPIs prediction. This idea is inspired from application of a VAE to chemical design using the latent space \cite{Gomez}. 
Finally, we perform a multi-objective optimization (MOO), see e.g. \cite{996017}, using the meta-model in the latent space for contradicting KPIs to obtain an improved machine design.

This paper is structured as follows: we discuss in the following section the dataset, its notation and the methodology. The \autoref{sec:NAT} addresses the network architecture and its training. In \autoref{sec:NR} a real-world example is demonstrated and finally the paper is concluded in \autoref{sec:conclusion}.

\begin{figure}
 	\centering
 	\input{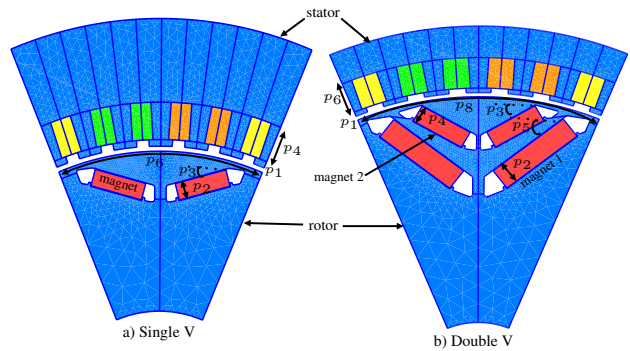}
 	\vspace{-3mm}
    \caption{Different machine topologies for PMSM}
    \label{fig:TOP}
\end{figure}

\section{Dataset and methodology}\label{sec:DM}
Our procedure for data generation is based on real-world industrial simulation workflow discussed in \cite{9333549}. \autoref{fig:TOP} illustrates modern-day machine designs. We use a large number of conventional magneto-static FE simulations, see e.g. \cite{Salon_1995aa}, and consider $k=1,\ldots,N$ different machine designs described by parameterizations $\mathbf{p}_k\in\mathbb{P}_{k}\subset\mathbb{R}^{n_k}$.
We assume that the KPIs are identical for each topology and 
are denoted by the vector $\mathbf{\kpi}_k(\mathbf{p}_{k})$. The entries may for example correspond to material cost (Euro), maximum torque-ripple at limiting curve (Nm), maximum torque (Nm) or maximum power (W). 
An integrated design space $\mathbb{P}\subset\mathbb{R}^{n}$ with $n=1+n_1+\ldots+n_{N}$ is constructed from the individual parameter spaces of the topologies, i.e., we obtain for the $j$-th sample  ($j=1,\ldots,S_{\textrm{tot}}$) if it belongs of topology $k$
$$
\mathbf{p}^{(j)}=[k,\mathbf{0},\ldots,\mathbf{0},\mathbf{p}_{k}^{(j)},\mathbf{0},\ldots ,\mathbf{0}]
\quad\text{and}\quad
\mathbf{y}^{(j)}=\mathbf{\kpi}_k(\textbf{p}_{k}^{(j)}).
$$
Finally, the complete input dataset can be written as
\begin{align}
\mathbf{P} :=\Big\{ \mathbf{p}^{(j)} \;\Big|\; \text{for } j=1,\ldots,S_{\textrm{tot}}\, \Big\}
\end{align}
which represent all the samples for all topologies. The VAE provides a probabilistic way for describing complex high-dimensional data in a hidden space and enables sampling of new data in the original space from it. It consists of an encoder and a decoder, each of which can be represented by neural networks. In this work, we introduced MLP along with usual VAE structure in the latent space for KPIs prediction. The model training is carried out as proposed in~\cite{MAL-056}. 

It is assumed that all input samples in $\mathbf{P}$ are generated by the $n$-dimensional random variable $\mathbf{p}$ and can be represented through unobserved variables $\mathbf{z}$ from a latent space of dimension $m \leq n$.
The encoder network approximates the conditional distribution $\mathcal{P}(\mathbf{z}|\mathbf{p})$. Any distribution can be chosen but the multivariate Gaussian with diagonal covariance is commonly used in practice. We follow this, i.e. $\mathbf{z}$ is assumed to be standard normal distribution. The encoder network determines the output parameters  mean ($\bm{\upsilon}$) and diagonal entries of covariance matrix ($\bm{\sigma}$) as a vector with a dimension $m$. It can be  described as
\begin{align}
        \label{eq:encoder}
        ( \bm{\upsilon},\bm{\sigma}) := \mathbf{e}_{\theta}(\mathbf{p})
\end{align}
where $\theta$ are trainable network parameters of the encoder network $\mathbf{e}_{\theta}$.
The latent vector $\mathbf{z}$ is sampled with the reparameterization trick (for efficient gradient training) by adding a noise vector $\bm{\varepsilon}$ of dimension $m$ as described in \cite{kingma2014auto}, i.e.,
\begin{align}
    \mathbf{z} = \bm{\upsilon} + \bm{\sigma} \odot   \bm{\varepsilon}
\end{align}
where $\bm{\varepsilon} \sim \mathcal{N}(0, \mathbf{I})$, and $\odot$ is the element-wise dot product. This latent variable is fed as input to the decoder network $\mathbf{d}_{\phi}$. It models the conditional distribution $\mathcal{P}(\mathbf{p}|\mathbf{z})$ where $\phi$ are the trainable parameters of the decoder, i.e.,
\begin{align}
\hat{\mathbf{p}} := \mathbf{d}_{\phi}(\mathbf{z})
\end{align}
The decoder $\mathbf{d}_{\phi}$ works as a design predictor while optimizing KPIs in the latent space.
The MLP is being trained alongside the continuous latent space to predict the KPIs
\begin{align}
    \hat {\mathbf{y}} := \mathbf{k}_{\psi}(\mathbf{z})
\end{align}
where $\psi$ are the trainable network parameters and $\hat {\mathbf{y}}$ is the predicted vector of KPIs. 
The training process is carried out with standard back-propagation. The objective of the training process is to improve encoding, reconstruction and prediction process by concurrently optimizing the model parameters $\theta,\phi, \psi$. The training loss includes three terms: parameter reconstruction loss ($\ell_{2}$-norm), Kullback–Leibler (KL) divergence (regularization) and loss ($\ell_{2}$-norm) for the KPIs prediction (MLP). Since we have parameter-based input data, the $\ell_{2}$-norm is a reasonable choice. The training loss
\begin{align}
\mathcal{L}(\theta,\phi,\psi;(\mathbf{p}^{(j)},\mathbf{y}^{(j)})) 
    &=  \left \|\mathbf{p}^{(j)} - \hat{\mathbf{p}}^{(j)} \right \|^{2}
    \!\!\!+\! \left \|\mathbf{y}^{(j)} -  \hat{\mathbf{y}}^{(j)} \right \|^{2}\!\!\!\nonumber\\ 
    &+ \mathcal{D}_{\mathrm{KL}}\Big(\mathcal{P}(\mathbf{z}^{(j)}|\mathbf{p}^{(j)},\theta) \;||\; \mathbf{z} \sim \mathcal{N}(0,\mathbf{I})\Big)
\end{align}

is defined with respect to model parameters, input observation $\mathbf{p}^{(j)}$ and ground truth KPIs $\mathbf{y}^{(j)}$. The KL divergence $\mathcal{D}_{\textrm{KL}}$ ensures that the approximation of the encoder distribution is close to the defined prior distribution over the latent variables.

\begin{figure}[h!]
	\centering
	\tikzset{every picture/.style={line width=0.75pt}} 

\begin{tikzpicture}[x=0.75pt,y=0.75pt,yscale=-1,xscale=1]

\draw   (96.24,17.3) -- (154.87,34.39) -- (155.13,69.3) -- (96.78,87.29) -- cycle ;
\draw   (293.7,87) -- (234.2,69.15) -- (234.2,34.85) -- (293.7,17) -- cycle ;
\draw   (241.34,94.3) -- (299.97,111.39) -- (300.23,146.3) -- (241.88,164.29) -- cycle ;
\draw  [fill={rgb, 255:red, 155; green, 155; blue, 155 }  ,fill opacity=1 ] (87.8,16.5) -- (94.5,16.5) -- (94.5,87.8) -- (87.8,87.8) -- cycle ;
\draw  [fill={rgb, 255:red, 155; green, 155; blue, 155 }  ,fill opacity=1 ] (295.6,16.5) -- (303.1,16.5) -- (303.1,88.25) -- (295.6,88.25) -- cycle ;
\draw  [fill={rgb, 255:red, 155; green, 155; blue, 155 }  ,fill opacity=1 ] (222.2,35.4) -- (231.7,35.4) -- (231.7,69.65) -- (222.2,69.65) -- cycle ;
\draw  [fill={rgb, 255:red, 155; green, 155; blue, 155 }  ,fill opacity=1 ] (301.38,111.5) -- (306.42,111.5) -- (306.42,146.75) -- (301.38,146.75) -- cycle ;
\draw  [fill={rgb, 255:red, 155; green, 155; blue, 155 }  ,fill opacity=1 ] (157.51,35) -- (172.59,35) -- (172.59,69.25) -- (157.51,69.25) -- cycle ;


\draw    (172.6,47.8) -- (191.2,47.8) ;
\draw [shift={(194.2,47.8)}, rotate = 180] [fill={rgb, 255:red, 0; green, 0; blue, 0 }  ][line width=0.08]  [draw opacity=0] (3.57,-1.72) -- (0,0) -- (3.57,1.72) -- cycle    ;
\draw    (157,52.1) -- (172.2,51.8) ;
\draw    (172.6,65) -- (191.6,65) ;
\draw [shift={(194.6,65)}, rotate = 182.08] [fill={rgb, 255:red, 0; green, 0; blue, 0 }  ][line width=0.08]  [draw opacity=0] (3.57,-1.72) -- (0,0) -- (3.57,1.72) -- cycle    ;
\draw    (198.5,83.85) -- (198.5,72.35) ;
\draw [shift={(198.5,69.35)}, rotate = 450] [fill={rgb, 255:red, 0; green, 0; blue, 0 }  ][line width=0.08]  [draw opacity=0] (3.57,-1.72) -- (0,0) -- (3.57,1.72) -- cycle    ;
\draw    (198.5,61.25) -- (198.5,52.25) ;
\draw [shift={(198.5,51.25)}, rotate = 450] [fill={rgb, 255:red, 0; green, 0; blue, 0 }  ][line width=0.08]  [draw opacity=0] (3.57,-1.72) -- (0,0) -- (3.57,1.72) -- cycle    ;
\draw    (202,47.8) -- (218.4,47.8) ;
\draw [shift={(221.4,47.8)}, rotate = 180] [fill={rgb, 255:red, 0; green, 0; blue, 0 }  ][line width=0.08]  [draw opacity=0] (3.57,-1.72) -- (0,0) -- (3.57,1.72) -- cycle    ;
\draw    (229.5,69.85) -- (229.5,150.35) ;
\draw    (229.8,149.85) -- (238.3,149.85) ;
\draw [shift={(241.3,149.85)}, rotate = 180] [fill={rgb, 255:red, 0; green, 0; blue, 0 }  ][line width=0.08]  [draw opacity=0] (3.57,-1.72) -- (0,0) -- (3.57,1.72) -- cycle    ;
\draw   (229.3,168.85) -- (288.8,168.85) -- (288.8,183.6) -- (229.3,183.6) -- cycle ;
\draw    (225,176.6) -- (225.1,72.65) ;
\draw [shift={(225.1,69.65)}, rotate = 450.05] [fill={rgb, 255:red, 0; green, 0; blue, 0 }  ][line width=0.08]  [draw opacity=0] (3.57,-1.72) -- (0,0) -- (3.57,1.72) -- cycle    ;
\draw    (225,176.2) -- (229.8,175.8) ;
\draw    (303.8,147.4) -- (304.2,177.4) ;
\draw    (304.7,177.15) -- (291.2,177.03) ;
\draw [shift={(288.2,177)}, rotate = 360.52] [fill={rgb, 255:red, 0; green, 0; blue, 0 }  ][line width=0.08]  [draw opacity=0] (3.57,-1.72) -- (0,0) -- (3.57,1.72) -- cycle    ;
\draw    (226.5,27.95) -- (226.5,32.95) ;
\draw [shift={(226.5,35.95)}, rotate = 270] [fill={rgb, 255:red, 0; green, 0; blue, 0 }  ][line width=0.08]  [draw opacity=0] (3.57,-1.72) -- (0,0) -- (3.57,1.72) -- cycle    ;

\draw (86.2,88.9) node [anchor=north west][inner sep=0.75pt]  [font=\small]  {$\mathbf{p}$};
\draw (294.6,88.9) node [anchor=north west][inner sep=0.75pt]  [font=\small]  {$\hat{\mathbf{p}}$};
\draw (116.5,41.4) node [anchor=north west][inner sep=0.75pt]    {$\mathbf{e}_{\theta }$};
\draw (250.6,41.4) node [anchor=north west][inner sep=0.75pt]    {$\mathbf{d}_{\phi }$};
\draw (248.7,122.6) node [anchor=north west][inner sep=0.75pt]    {$\mathbf{k}_{\psi }$};
\draw (307.3,111.8) node [anchor=north west][inner sep=0.75pt] [font=\small]   {$\hat{\mathbf{y}}$};
\draw (231.8,170.1) node [anchor=north west][inner sep=0.75pt]  [font=\scriptsize] [align=left] {optimization};
\draw (192,59.4) node [anchor=north west][inner sep=0.75pt]    {$\odot $};
\draw (192,42) node [anchor=north west][inner sep=0.75pt]  [font=\small,rotate=-359.81]  {$\oplus$};
\draw (160.8,57.6) node [anchor=north west][inner sep=0.75pt]  [font=\small]  {$\bm{\sigma }$};
\draw (160.4,39.9) node [anchor=north west][inner sep=0.75pt]  [font=\small]  {$\bm{\upsilon }$};
\draw (222.1,46.3) node [anchor=north west][inner sep=0.75pt]  [font=\small]  {$\mathbf{z}$};
\draw (161,84) node [anchor=north west][inner sep=0.75pt]  [font=\small]  {$\bm{\varepsilon } \sim \mathcal{N} (0,\mathbf{I} )$};
\draw (197.5,16) node [anchor=north west][inner sep=0.75pt]  [font=\small] [align=left] {latent vector};
\draw (174.2,30.5) node [anchor=north west][inner sep=0.75pt]  [font=\small] [align=left] {sampling};
\draw (108.1,31.3) node [anchor=north west][inner sep=0.75pt]  [font=\small] [align=left] {encoder};
\draw (248,31.3) node [anchor=north west][inner sep=0.75pt]  [font=\small] [align=left] {decoder};
\draw (242.8,111.2) node [anchor=north west][inner sep=0.75pt]  [font=\small] [align=left] {KPIs pred.};

\end{tikzpicture}
	\vspace{-3mm}
	\caption{Proposed VAE-based workflow}
	\label{fig:vae}
\end{figure}
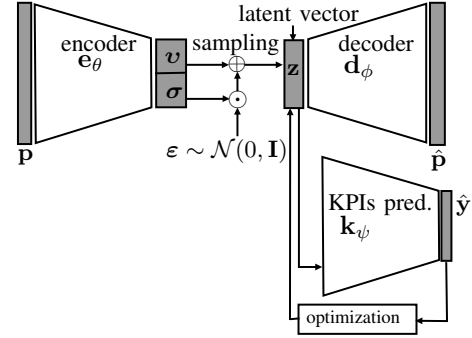

\begin{table}[h!]
\caption{KPIs summary and evaluation}
\vspace{-3mm}
\label{tab:KPIs}
\resizebox{\linewidth}{!}{%
\begin{tabular}{|c|c|c|c|c|c|c|}
\hline
\multirow{2}{*}{} & \multirow{2}{*}{\textbf{KPIs}} & \multirow{2}{*}{\textbf{Unit}} & \multicolumn{4}{c|}{\textbf{Prediction accuracy}} \\ \cline{4-7} 
   &                       &      & \textbf{MAE} & \textbf{RMSE} & \textbf{PCC} & \textbf{MRE(\%)} \\ \hline
$\kpi_{ 1}$ & Maximum torque        & Nm   & 2.45         & 3.15          & 0.99         & 0.55             \\ \hline
$\kpi_{ 2}$ & Maximum power         & KW   & 1.64         & 2.2           & 0.99         & 0.49             \\ \hline
$\kpi_{ 3}$ & Maximum torque ripple & Nm   & 3.52         & 5.01          & 0.99         & 4.05             \\ \hline
$\kpi_{ 4}$ & Material cost         & Euro & 1.43         & 1.8           & 0.99         & 0.64             \\ \hline
\end{tabular}%
}
\end{table}

\begin{table}[]
\caption{SV parameters detail and evaluation}
\vspace{-3mm}
\label{tab:svp}
\resizebox{\linewidth}{!}{%
\begin{tabular}{|c|c|c|c|c|c|c|c|c|}
\hline
\multirow{2}{*}{\textbf{}} &
  \multirow{2}{*}{\textbf{Parameters}} &
  \multirow{2}{*}{\textbf{Min.}} &
  \multirow{2}{*}{\textbf{Max.}} &
  \multirow{2}{*}{\textbf{Unit}} &
  \multicolumn{4}{c|}{\textbf{Reconstruction accuracy}} \\ \cline{6-9} 
   &                             &     &     &     & \textbf{MAE} & \textbf{RMSE} & \textbf{PCC} & \textbf{MRE(\%)} \\ \hline
$p_{1}$ & Air gap                     & 0.8 & 1.8 & mm  & 0.009        & 0.01          & 0.99         & 0.65             \\ \hline
$p_{2}$ & Height of magnet            & 4.5 & 6.5 & mm  & 0.01         & 0.012         & 1            & 0.18             \\ \hline
$p_{3}$ & Inclination angle of magnet & 14  & 36  & deg & 0.146        & 0.175         & 0.99         & 0.71             \\ \hline
$p_{4}$ & Stator tooth height         & 12  & 20  & mm  & 0.006        & 0.006         & 0.99         & 0.39             \\ \hline
$p_{5}$ & Iron length                 & 120 & 160 & mm  & 0.036        & 0.047         & 1            & 0.23             \\ \hline
$p_{6}$ & Rotor outer diameter        & 150 & 180 & mm  & 0.017        & 0.019         & 1            & 0.28             \\ \hline
\end{tabular}%
}
\end{table}

\begin{table}[]
\caption{DV parameters detail and evaluation}
\vspace{-3mm}
\label{tab:dvp}
\resizebox{\linewidth}{!}{%
\begin{tabular}{|c|c|c|c|c|c|c|c|c|}
\hline
\multirow{2}{*}{} &
  \multirow{2}{*}{\textbf{Parameters}} &
  \multirow{2}{*}{\textbf{Min.}} &
  \multirow{2}{*}{\textbf{Max.}} &
  \multirow{2}{*}{\textbf{Unit}} &
  \multicolumn{4}{c|}{\textbf{Reconstruction accuracy}} \\ \cline{6-9} 
   &                               &     &     &     & \textbf{MAE} & \textbf{RMSE} & \textbf{PCC} & \textbf{MRE(\%)} \\ \hline
$p_{1}$ & Air gap                       & 0.8 & 1.8 & mm  & 0.004        & 0.006         & 1            & 0.37             \\ \hline
$p_{2}$ & Height of magnet 1            & 4.5 & 6.5 & mm  & 0.022        & 0.024         & 0.99         & 0.39             \\ \hline
$p_{3}$ & Inclination angle of magnet 1 & 20  & 40  & deg & 0.167        & 0.191         & 0.99         & 0.56             \\ \hline
$p_{4}$ & Height of magnet 2            & 3.7 & 5.6 & mm  & 0.009        & 0.011         & 1            & 0.19             \\ \hline
$p_{5}$ & Inclination angle of magnet 2 & 18  & 35  & deg & 0.173        & 0.188         & 0.99         & 0.67             \\ \hline
$p_{6}$ & Stator tooth height           & 10  & 24  & mm  & 0.002        & 0.003         & 0.99         & 0.19             \\ \hline
$p_{7}$ & Iron length                   & 120 & 160 & mm  & 0.017        & 0.021         & 1            & 0.31             \\ \hline
$p_{8}$ & Rotor outer diameter          & 150 & 180 & mm  & 0.01         & 0.012         & 1            & 0.53             \\ \hline
\end{tabular}%
}
\end{table}
\section{Network architecture and Training}\label{sec:NAT}
\subsection{Network architecture}
\begin{figure}
 	 \centering
 	 \tikzstyle{block_input_layers} = [rectangle, draw, text width=7em, text centered, rounded corners, minimum height=1em,font=\notsotiny ]
\tikzstyle{block_output_layers} = [rectangle, draw, fill = magenta!30, text width=2.85em, text centered, minimum height=1em,font=\tiny ]
\tikzstyle{block_output_layers2} = [rectangle, draw, fill = brown!30, 	text width=7em, text centered,  minimum height=1em,font=\notsotiny ]
\tikzstyle{block_intermediate_layers} = [rectangle, draw,     fill = blue!30, 	text width=7em, text centered, minimum height=1em,font=\notsotiny ]
\tikzstyle{dotted_block} = [draw=black!, line width=0.5pt, dash pattern=on 1pt off 4pt on 6pt off 4pt,inner ysep=0.8mm,inner xsep=0.8mm, rectangle, rounded corners]
\tikzstyle{line} = [draw, -latex']

\begin{tikzpicture}[node distance = 0.50em and 0.7em, auto]

	\node [block_input_layers] (E1) {input vector($\mathbf{p}$): $32 \times 1$};
	\node [block_intermediate_layers, below =of E1] (E2) {$32\times5$ Conv1D, stride 1 + tanh};
	\node [block_intermediate_layers, below =of E2] (E3) {$32\times10$ Conv1D, stride 1 + tanh};
	\node [block_intermediate_layers, below =of E3] (E4) {$32\times20$ Conv1D, stride 1 + tanh};
    \node [block_intermediate_layers, below =of E4] (E5) {640 flatten layer };
    \node [block_intermediate_layers, below =of E5] (E6) {400 \ Dense, tanh};
    \node[block_output_layers, below right= 0.7em and -1.25cm of E6](out1){$\bm{\upsilon}$: 19 Dense};

    \node[block_output_layers, left =  of out1](out2){$\bm{\sigma}$: 19 Dense};

	\path [line] (E1) -- (E2); 
	\path [line] (E2) -- (E3); 
	\path [line] (E3) -- (E4); 
	\path [line] (E4) -- (E5); 
	\path [line] (E5) -- (E6); 

	\node [block_input_layers, right =of E1] (D1) {latent vector ($\mathbf{z}$): $19 \times 1$};
	\node [block_intermediate_layers, below =of D1] (D2) {640 \ Dense, tanh};
	\node [block_intermediate_layers, below =of D2] (D3) {400 \ Dense, tanh};
	\node [block_intermediate_layers, below =of D3] (D4) {$32\times20$ Cov1DT, stride 1 + tanh};
    \node [block_intermediate_layers, below =of D4] (D5) {$32\times10$ Cov1DT, stride 1 + tanh};
    \node [block_intermediate_layers, below =of D5] (D6) {$32\times5$ Cov1DT, stride 1 + tanh};
    \node [block_intermediate_layers, below =of D6] (D7) {$32\times1$ Cov1DT, stride 1 + linear};
    \node [block_output_layers2, below =of D7] (D8) {pred.parameters ($\hat{\mathbf{p}}$): 32};
    \node [block_output_layers2, left =of D8] (E8) {latent vector ($\mathbf{z}$): $19 \times 1$};
   
    \node(v0) at ([xshift=0.73cm, yshift=-0.4em]E8.north){};
    \draw [line] (out1.south) -- (v0);
     \node(v1) at ([xshift=-0.76cm, yshift=-0.4em]E8.north){};
    \draw [line] (out2.south) -- (v1);
    
    \node(v2) at ([xshift=0.76cm, yshift=-0.88em]E6.south){};
    \draw [line] (E6) -- (v2);
     \node(v3) at ([xshift=-0.76cm, yshift=-0.88em]E6.south){};
    \draw [line] (E6) -- (v3);
    
	\path [line] (D1) -- (D2); 
	\path [line] (D2) -- (D3); 
	\path [line] (D3) -- (D4); 
	\path [line] (D4) -- (D5); 
	\path [line] (D5) -- (D6); 
	\path [line] (D6) -- (D7); 
	\path [line] (D7) -- (D8); 
    \node [dotted_block, fit = (D2) (D7)] (de) {};
    \node [dotted_block, fit = (E2) (out1) (out2)] (ee) {};

    \node [block_input_layers, right =of D1] (K1) {latent vector ($\mathbf{z}$): $19 \times 1$};
	\node [block_intermediate_layers, below =of K1] (K2) {250 \ Dense, softplus};
	\node [block_intermediate_layers, below =of K2] (K3) {250 \ Dense, softplus};
	\node [block_intermediate_layers, below =of K3] (K4) {180 \ Dense, softplus};
    \node [block_intermediate_layers, below =of K4] (K5) {80 \ Dense, softplus};
    \node [block_intermediate_layers, below =of K5] (K6) {50 \ Dense, softplus};
    \node [block_output_layers2, below =of K6] (K7) {output KPIs ($\hat{\mathbf{y}}$): 4};

	\path [line] (K1) -- (K2); 
	\path [line] (K2) -- (K3); 
	\path [line] (K3) -- (K4); 
	\path [line] (K4) -- (K5); 
	\path [line] (K5) -- (K6); 
	\path [line] (K6) -- (K7); 
    \node [dotted_block, fit = (K2) (K6)] (ke) {};

    \node(v4) at ([xshift=-0.0cm, yshift=-0.6em]E8.south) [font=\scriptsize]{a) Encoder};
    \node(v5) at ([xshift=-0.0cm, yshift=-0.6em]D8.south) [font=\scriptsize]{b) Decoder};
    \node(v6) at ([xshift=-0.0cm, yshift=-0.6em]K7.south)[font=\scriptsize]{c) KPIs Predictor};

\end{tikzpicture}
 	  \vspace{-4mm}
     \caption{Network structure}
     \label{fig:nat}
\end{figure}
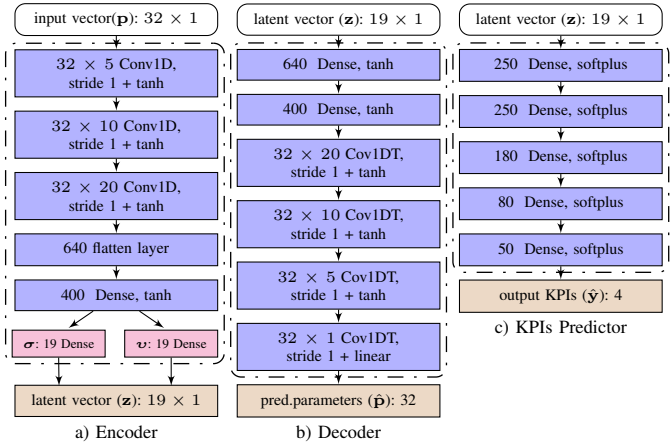
\autoref{fig:vae} gives gist of complete workflow.
For our demonstration, we created datasets for $N=2$ topologies: Single-V (SV) and Double-V (DV). \autoref{fig:TOP} depicts representative samples for both topologies. The SV topology has $n_1 = 13$ parameters, while the DV topology has $n_2 = 18$. These numbers are determined at random based on experience. As a result, the integrated space becomes $n = 32$ with an added topology indicator parameter ($k$). 

The network is split into three parts: encoder ($\mathbf{e}_{\theta }$), decoder ($\mathbf{d}_{\phi}$) and KPIs predictor ($\mathbf{k}_{\psi}$). The architecture and its hyperparameters are derived through trial and error after evaluating roughly twenty five different configurations, in detail:

\begin{itemize}
     \item \textit{Encoder}: The encoder is interpreted as an inference network to approximate posterior for the latent space. It consists of three $1-d$ convolutional layers (to determine effectively whether the design is SV or DV using the relationship between topology indicator and parameters in the integrated design space), one flatten layer followed by a dense layer and three output layers. The output layers which have the size of the latent space encompasses the distribution parameters mean ($\bm{\upsilon}$) and variance ($\bm{\sigma}$) that sample latent vector ($\mathbf{z}$) with the sampling layer. 
  \item \textit{Decoder}: The decoder (generative model) predicts the design parameters for the input latent vector ($\mathbf{z}$). The network structure, along with hyperparameters, is detailed in \autoref{fig:nat}. The network hyperparameters such as number of filters, filter size, stride, neurons per layer and activation function remains same as for the encoder layers except for a linear activation function in the output layer.

\item \textit{KPIs predictor}: The MLP is trained to predict KPIs in the latent space. The structure of MLP is depicted in \autoref{fig:nat}. It has five dense layers and an output layer of a size that corresponds to the number of KPIs to be predicted.
\end{itemize}

\subsection{Training}

\begin{figure}
 	\centering
 	\input{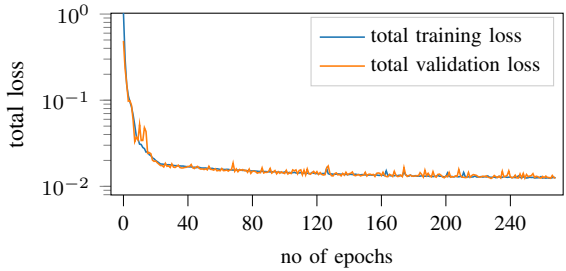}
 	\vspace{-5mm}
    \caption{Training and validation loss curves}
    \label{fig:val_loss}
\end{figure}

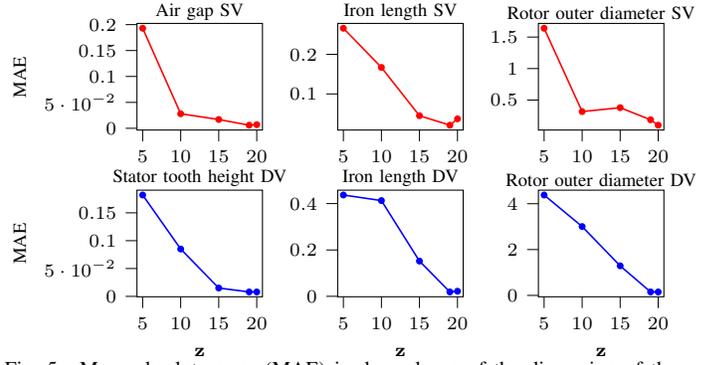
\begin{figure}
        \centering
\begin{tikzpicture}
\pgfplotsset{every axis title/.append style={at={(0.5,0.8)}}}
\tikzstyle{every node}=[font=\scriptsize]
\definecolor{color0}{rgb}{0.12156862745098,0.466666666666667,0.705882352941177}
\definecolor{color1}{rgb}{1,0.498039215686275,0.0549019607843137}

\begin{groupplot}[group style={group size=3 by 2,
                 horizontal sep = 1 cm, 
            vertical sep = 0.8cm}, 
        width = 3.25 cm, 
        height = 3 cm,] 
\nextgroupplot[
tick align=outside,
tick pos=left,
title={Air gap SV},
xmin=4.25, xmax=20.75,
xtick style={color=black},
y grid style={white!69.0196078431373!black},
ylabel={MAE},
ymin=-0.00335000013001263, ymax=0.202350003877655,
ytick style={color=black}
]
\addplot [semithick, red, mark=*, mark size=1, mark options={solid}]
table {%
5 0.193000003695488
10 0.0280000008642673
15 0.017000000923872
19 0.00600000005215406
20 0.00700000021606684
};

\nextgroupplot[
tick align=outside,
tick pos=left,
title={Iron length SV},
x grid style={white!69.0196078431373!black},
xmin=4.25, xmax=20.75,
xtick style={color=black},
y grid style={white!69.0196078431373!black},
ymin=0.00874999957159162, ymax=0.278250002767891,
ytick style={color=black}
]
\addplot [semithick, red, mark=*, mark size=1, mark options={solid}]
table {%
5 0.266000002622604
10 0.166999995708466
15 0.0450000017881393
19 0.0209999997168779
20 0.0370000004768372
};

\nextgroupplot[
tick align=outside,
tick pos=left,
title={Rotor outer diameter SV},
x grid style={white!69.0196078431373!black},
xmin=4.25, xmax=20.75,
xtick style={color=black},
y grid style={white!69.0196078431373!black},
ymin=0.0240500044077635, ymax=1.7169499848038,
ytick style={color=black}
]
\addplot [semithick, red, mark=*, mark size=1, mark options={solid}]
table {%
5 1.63999998569489
10 0.31700000166893
15 0.379000008106232
19 0.187000006437302
20 0.101000003516674
};

\nextgroupplot[
tick align=outside,
tick pos=left,
title={Stator tooth height DV},
x grid style={white!69.0196078431373!black},
xlabel={\(\displaystyle \mathbf{z}\)},
xmin=4.25, xmax=20.75,
xtick style={color=black},
y grid style={white!69.0196078431373!black},
ylabel={MAE},
ymin=-0.000699999416247011, ymax=0.190699996100739,
ytick style={color=black}
]
\addplot [semithick, blue, mark=*, mark size=1, mark options={solid}]
table {%
5 0.181999996304512
10 0.0850000008940697
15 0.0149999996647239
19 0.00800000037997961
20 0.00800000037997961
};

\nextgroupplot[
tick align=outside,
tick pos=left,
title={Iron length DV},
x grid style={white!69.0196078431373!black},
xlabel={\(\displaystyle \mathbf{z}\)},
xmin=4.25, xmax=20.75,
xtick style={color=black},
y grid style={white!69.0196078431373!black},
ymin=-0.00190000096336007, ymax=0.457900006789714,
ytick style={color=black}
]
\addplot [semithick, blue, mark=*, mark size=1, mark options={solid}]
table {%
5 0.437000006437302
10 0.412999987602234
15 0.151999995112419
19 0.0189999993890524
20 0.0219999998807907
};

\nextgroupplot[
tick align=outside,
tick pos=left,
title={Rotor outer diameter DV},
x grid style={white!69.0196078431373!black},
xlabel={\(\displaystyle \mathbf{z}\)},
xmin=4.25, xmax=20.75,
xtick style={color=black},
y grid style={white!69.0196078431373!black},
ymin=-0.0604500129818916, ymax=4.59145012050867,
ytick style={color=black}
]
\addplot [semithick, blue, mark=*, mark size=1, mark options={solid}]
table {%
5 4.38000011444092
10 3.00399994850159
15 1.28900003433228
19 0.151999995112419
20 0.150999993085861
};
\end{groupplot}

\end{tikzpicture}
 	     \vspace{-9mm}
        \caption{Mean absolute error (MAE) in dependence of the dimension of the latent space}
        \label{fig:latent_space}
\end{figure}

The SV and DV designs have $S_1 = 14854$ and $S_2 = 13424$ samples, respectively, such that the total number is $S_{\mathrm{tot}} = 28278$. The whole network (encoder, decoder and MLP) is trained concurrently with $\sim 90 \%$ of $S_{\mathrm{tot}}$ in a combined parameter space. Around, $5\%$ of the total samples are kept for both validation and testing. \autoref{fig:val_loss} exhibits training and validation curves. The training hyperparameters are obtained by trial and error. It includes the optimizer (Adam \cite{kingma2015adam}), learning rate ($0.001$--$0.0001$), epochs ($300$) with validation patience ($10$), loss ($\ell_{2}$-norm and KL-divergence), latent space dimension ($19$), batch size ($40$). The entire model was trained on a Quadro M2000M GPU and took $\sim 12$ minutes to complete model training. Decoder and MLP analyze new designs in roughly $\sim 3-4$\,ms/sample, that is much faster than the FE simulation, which takes around  $\sim 4-6$\,h/sample on a single core CPU. It was observed that for a reasonable parameter reconstruction, the latent space dimension should be set larger than the maximum number of topology parameters, i.e.$\max_k (n_k)$. Since the DV topology has the maximal number of parameters ($n_{2} = 18$), the latent space dimension is tuned to $19$. 
The evaluation for three input parameters of each topology with latent dimensions ranging from $5$ to $20$ is shown in \autoref{fig:latent_space}.
\section{Numerical results}\label{sec:NR}
\begin{figure}
    \centering 
    \includegraphics[width = 0.65\columnwidth]{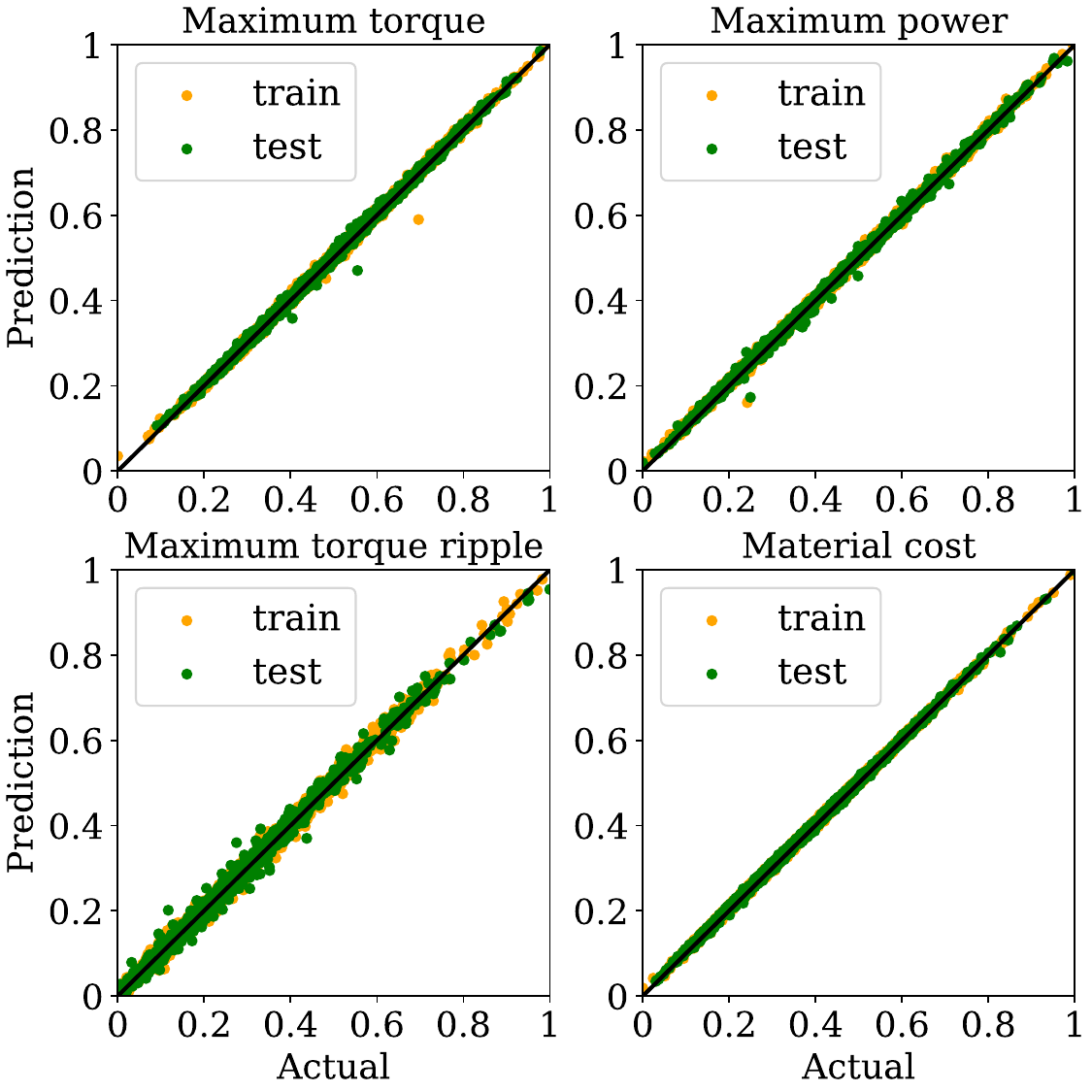}
    \vspace{-4mm}
    \caption{Predictions of the KPIs over test samples}
    \label{fig:kpi_pred}
\end{figure}

\begin{figure}
    \centering                  
    \includegraphics[width=0.90\columnwidth]{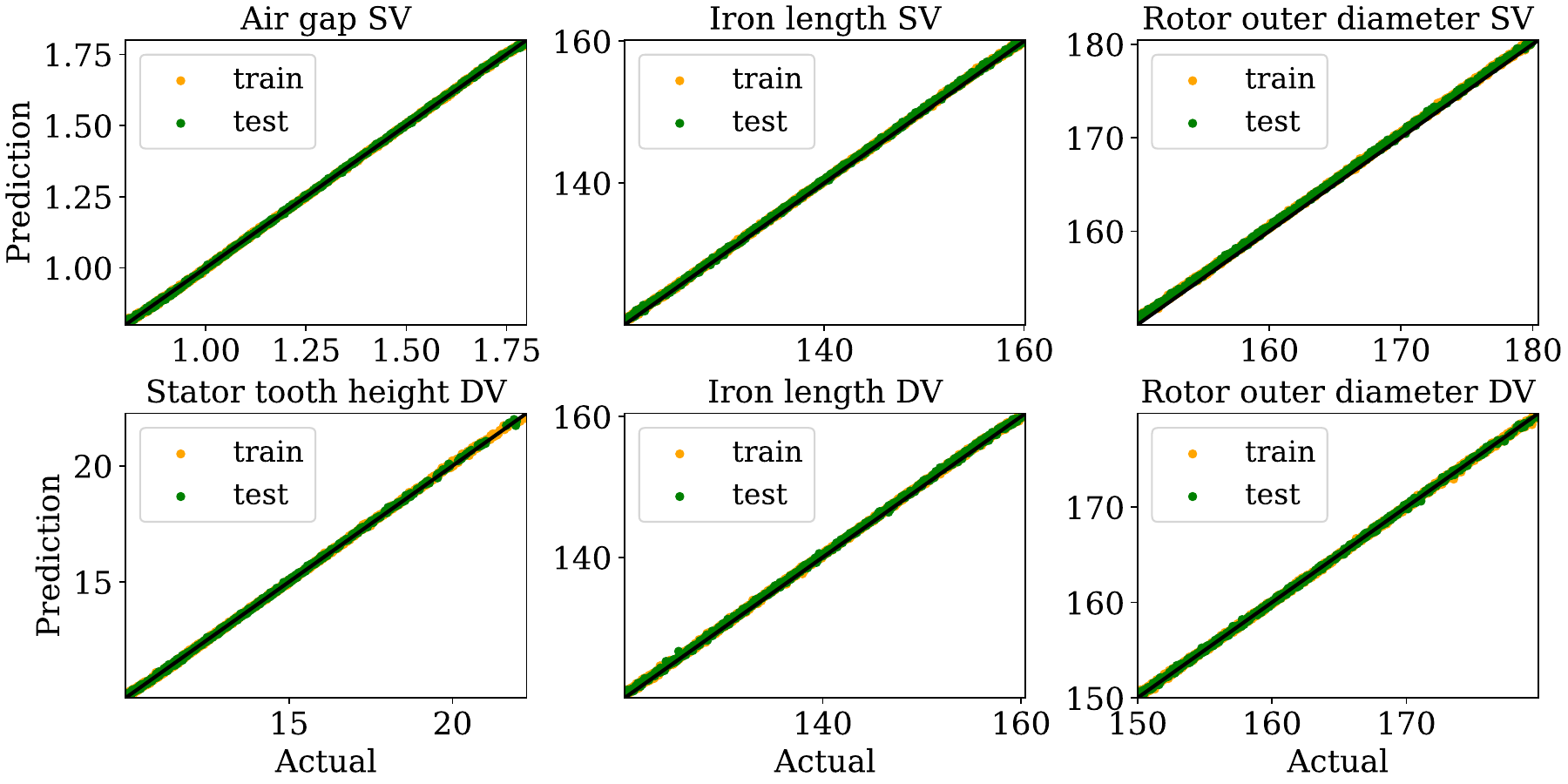}
    \vspace{-4mm}
    \caption{Parameters prediction plot over test samples}
    \label{fig:param_plot}
\end{figure}

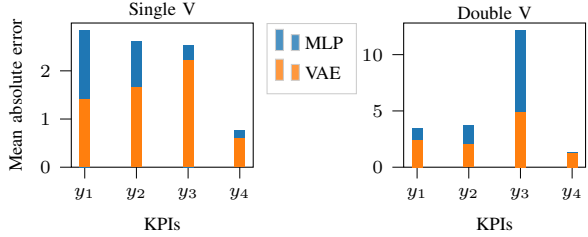
\begin{figure}
        \centering
\begin{tikzpicture}

\definecolor{color0}{rgb}{0.12156862745098,0.466666666666667,0.705882352941177}
\definecolor{color1}{rgb}{1,0.498039215686275,0.0549019607843137}
\tikzstyle{every node}=[font=\scriptsize]
\pgfplotsset{every axis title/.append style={at={(0.5,0.85)}}}
\begin{groupplot}[group style={group size=2 by 1,
                 horizontal sep = 2 cm, 
            vertical sep = 0.8cm}, 
        width = 4 cm, 
        height = 3.5 cm,]
\nextgroupplot[
legend cell align={left},
legend style={fill opacity=0.8, draw opacity=1, text opacity=1, at={(1.57,1)}, draw=white!80!black},
tick align=outside,
tick pos=left,
title={Single V},
x grid style={white!69.0196078431373!black},
xlabel={KPIs},
xmin=-0.26, xmax=3.26,
xtick style={color=black},
xtick={0,1,2,3},
xticklabels={
  \(\displaystyle y_{1}\),
  \(\displaystyle y_{2}\),
  \(\displaystyle y_{3}\),
  \(\displaystyle y_{4}\)
},
y grid style={white!69.0196078431373!black},
ylabel={Mean absolute error},
ymin=0, ymax=2.9925,
ytick style={color=black}
]
\draw[draw=none,fill=color0] (axis cs:-0.1,0) rectangle (axis cs:0.1,2.85);
\addlegendimage{ybar,ybar legend,draw=none,fill=color0}
\addlegendentry{MLP}
\draw[draw=none,fill=color0] (axis cs:0.9,0) rectangle (axis cs:1.1,2.61);
\draw[draw=none,fill=color0] (axis cs:1.9,0) rectangle (axis cs:2.1,2.54);
\draw[draw=none,fill=color0] (axis cs:2.9,0) rectangle (axis cs:3.1,0.76);
\draw[draw=none,fill=color1] (axis cs:-0.1,0)rectangle (axis cs:0.1,1.4);

\addlegendimage{ybar,ybar legend,draw=none,fill=color1}
\addlegendentry{VAE}

\draw[draw=none,fill=color1] (axis cs:0.9,0) rectangle (axis cs:1.1,1.66);
\draw[draw=none,fill=color1] (axis cs:1.9,0) rectangle (axis cs:2.1,2.21);
\draw[draw=none,fill=color1] (axis cs:2.9,0) rectangle (axis cs:3.1,0.6);
\nextgroupplot[
tick align=outside,
tick pos=left,
title={Double V},
x grid style={white!69.0196078431373!black},
xlabel={KPIs},
xmin=-0.26, xmax=3.26,
xtick style={color=black},
xtick={0,1,2,3},
xticklabels={
  \(\displaystyle y_{1}\),
  \(\displaystyle y_{2}\),
  \(\displaystyle y_{3}\),
  \(\displaystyle y_{4}\)
},
y grid style={white!69.0196078431373!black},
ymin=0, ymax=12.81,
ytick style={color=black}
]
\draw[draw=none,fill=color0] (axis cs:-0.1,0) rectangle (axis cs:0.1,3.48);
\draw[draw=none,fill=color0] (axis cs:0.9,0) rectangle (axis cs:1.1,3.69);
\draw[draw=none,fill=color0] (axis cs:1.9,0) rectangle (axis cs:2.1,12.2);
\draw[draw=none,fill=color0] (axis cs:2.9,0) rectangle (axis cs:3.1,1.32);
\draw[draw=none,fill=color1] (axis cs:-0.1,0) rectangle (axis cs:0.1,2.43);
\draw[draw=none,fill=color1] (axis cs:0.9,0) rectangle (axis cs:1.1,2.02);
\draw[draw=none,fill=color1] (axis cs:1.9,0) rectangle (axis cs:2.1,4.9);
\draw[draw=none,fill=color1] (axis cs:2.9,0) rectangle (axis cs:3.1,1.26);
\end{groupplot}

\end{tikzpicture}
        \vspace{-4mm}
        \caption{Comparison MLP vs VAE}
        \label{fig:cmp_dnn_vae}
\end{figure}%

\begin{table}[]
\caption{Design evaluation from pareto front}
\vspace{-3mm}
\label{tab:pareto_pred}
\resizebox{\linewidth}{!}{%
\begin{tabular}{|c|c|c|c|c|c|c|}
\hline
\multirow{2}{*}{\textbf{KPIs}} & \multicolumn{3}{c|}{\textbf{Design A (SV) }}                & \multicolumn{3}{c|}{\textbf{Design B (DV)}}                \\ \cline{2-7} 
                               & \textbf{FE simulation} & \textbf{Prediction} & \textbf{MRE(\%)} & \textbf{FE simulation} & \textbf{Prediction} & \textbf{MRE(\%)} \\ \hline
$\kpi_{ 1}$                             & 351.86                 & 346.79              & 1.44             & 489.36                 & 470.99              & 3.75             \\ \hline
$\kpi_{ 2}$                             & 284.34                 & 280.97              & 1.18             & 578.87                 & 600.96              & 3.8              \\ \hline
$\kpi_{ 3}$                             & 29.34                  & 31.92               & 8.79             & 232.95                 & 216.97              & 6.8              \\ \hline
$\kpi_{ 4}$                             & 131.8                  & 133.98              & 1.65             & 301.31                 & 308.71              & 2.4              \\ \hline
\end{tabular}%
}
\end{table}
 
The mean absolute error (MAE), root mean squared error (RMSE), Pearson correlation coefficient (PCC), and dimensionless mean relative error (MRE) are used as quality metrics for KPIs prediction and parameter reconstruction. \autoref{fig:kpi_pred} and \autoref{tab:KPIs} illustrates the predictions and evaluations for four KPIs over the test samples in the combined design space. Similarly, \autoref{tab:svp} and \autoref{tab:dvp} provide evaluation details for important parameters of each topology. Prediction plots for three geometry parameters to each topology are demonstrated in \autoref{fig:param_plot}. The \autoref{fig:cmp_dnn_vae} displays comparative evaluation of the prediction performance between MLP and VAE for each topology. The MLP for each topology is trained directly on the input parameters, as described in \cite{9333549}, using a supervised learning approach. The network configuration, training hyperparameters, and datasets (training, validation, and testing) are identical to the KPIs predictor for quantitative comparison. The VAE outperformed the MLP possibly due to a large number of training samples (through integrated parameter space) as compared to separately trained MLP for each topology, tuning of latent space dimension (slightly higher or the maximum number of topology parameters) that does better functional mapping between the latent input and output KPIs.
The proposed method is applied for optimization based on the $N=2$ topologies. 
The trained MLP is used as a meta-model for the latent space. 

\begin{figure}
 	\centering
 	\input{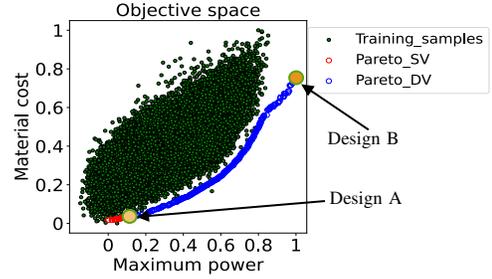}
 	\vspace{-5mm}
    \caption{Pareto-front for Material cost and Maximum power, where Pareto$\_$SV (red) refers to Single V and Pareto$\_$DV (blue) to Double V topologies.}
    \label{fig:moo}
\end{figure}

Finally, in the latent space, two opposing objectives, maximum power and material cost, are optimized for demonstration. We use the multi-objective genetic algorithm NSGA-\rom{2} \cite{996017}. The hyperparameter settings include sampling method (real random), random initialization, crossover, mutation, stopping criteria with maximum number of generations (100), population size (1000). The lower and upper bounds are determined using the mean values ($\bm{\upsilon}$) of model training samples. In order to improve  the design validation factor of the obtained designs in the pareto-front, the optimization process is constrained by the same bounds that were set during the data set generation. The optimization takes around $\sim 2$ hours. It can be seen from \autoref{fig:moo} that the pareto-front consists of samples (unseen during model training) from each of the two topologies, which are cost-effective and/or power-efficient. \autoref{fig:TOP} depicts two samples (Design A and Design B) from the pareto-front. For reference, the result of a FE simulation is compared with the prediction in \autoref{tab:pareto_pred}. The average MRE for KPIs prediction is less than $\sim 5\% $ for both samples. 

\section{Conclusion}\label{sec:conclusion}
This paper presents a VAE-based approach to derive a unified parameterization for multiple PMSM topologies with different parameterizations and predicting KPIs with high accuracy simultaneously. It enables concurrent parametric optimization in multi-topology scenarios. The numerical results demonstrate that latent space optimization improves over the training data and produces novel (unseen in the training data) designs. The latent space dimension was observed to be a key hyperparameter for parameter reconstruction precision. This research lays the groundwork for parameter-based latent space optimization in the domain of the electric machines. A possible next step would be to investigate the impact of other latent space priors (e.g., Gaussian mixture models). It will also be interesting to extend the approach by involving different machine types (e.g., induction machines and flux switching machines).
\bibliographystyle{ieeetran}
\bibliography{main_ref}
\end{document}